\documentclass{article}
\usepackage{spconf}
\usepackage{epsfig}
\usepackage{amsmath}
\usepackage{amssymb}
\usepackage{graphicx}
\usepackage{cite}
\usepackage{float}

\DeclareMathOperator*{\argmin}{argmin \; }

\DeclareMathOperator*{\minimize}{minimize \quad}
\DeclareMathOperator*{\subto}{subject\, to \quad}

\newcommand{\norm}[1]{ \left \Vert #1 \right \Vert }

\newcommand{\XSet}{\mathcal{X}}
\newcommand{\KSet}{\mathcal{K}}

\newcommand{\SSet}{\mathcal{S}}
\newcommand{\USet}{\mathcal{U}}

\newcommand{\Tr}{T}

\newcommand{\sepr}{\,|\,}

\newcommand{\Proj}{\mathcal{P}}

\newcommand{\ratnorms}{\ensuremath{\ell_1 / \ell_2} }

\newcommand{\bi}{\begin{itemize}}
\newcommand{\ei}{\end{itemize}}
\newcommand{\beq}{\begin{equation}}
\newcommand{\eeq}{\end{equation}}

\newcommand{\str}{\leftarrow}
\newcommand{\tru}{\text{true}}

\newcommand{\xf}{\mathbf{x}}
\newcommand{\yf}{\mathbf{y}}
\newcommand{\rf}{\mathbf{r}}
\newcommand{\nf}{\mathbf{n}}

\newcommand{\etal}{\emph{et al}}

\begin{document}
%\title{Proximity and Projection Operators for Hoyer's Robust Sparse Penalty}
%\title{Scale-Invariant Sparse Constrained Optimization and a Simple Blind Deconvolution Algorithm}
\title{Correcting Camera Shake by Incremental Sparse Approximation }%Edge Approximation}
\name{Paul Shearer, Anna C. Gilbert, Alfred O. Hero III}
\address{University of Michigan, Ann Arbor}
\date{\today}

\maketitle
\begin{abstract}
%are often afflicted by blur, so there is strong interest in automatically characterizing and correcting it%no fully satisfactory solution exists yet. 

The problem of deblurring an image when the blur kernel is unknown remains challenging after decades of work. Recently there has been rapid progress on correcting irregular blur patterns caused by camera shake, but there is still much room for improvement.
We propose a new blind deconvolution method using incremental sparse edge approximation to recover images blurred by camera shake. We estimate the blur kernel first from only the strongest edges in the image, then gradually refine this estimate by allowing for weaker and weaker edges. Our method competes with the benchmark deblurring performance of the state-of-the-art while being significantly faster and easier to generalize.%, and it can be adapted to nonuniform blur or non-Gaussian noise models. %, and other % and noise models. %non-Gaussian noise distributions, new kernel and image priors, and nonuniform blur.

\end{abstract}

%\abstract

\section{Introduction}
%In the blind deconvolution problem, we are given an image $y$ blurred by a kernel $k$ with additive noise $n$ and challenged to determine the latent sharp image $x$. 
In the problem of blind image deconvolution, we are given a blurry image $y$ and challenged to determine an estimate $x$ of the unknown sharp image $x^\tru$ without knowledge of the blur kernel $k^\tru$. In the simplest model of blur, $y$ is formed by convolving $x^\tru$ with  $k^\tru$ and adding noise $n$:
\beq
	y = k^\tru * x^\tru + n.
	\label{conv_model}
\eeq
This convolution model assumes spatially uniform blur, which is frequently violated due to slight camera rotations and out-of-plane effects  \cite{levin2009understanding}. Still, the uniform model works surprisingly well and methods for it can be extended to handle nonuniform blur \cite{cho2007removing,whyte2010nonuniform}.

Even with uniform blur by a single kernel, the blind deconvolution problem is highly underdetermined and additional assumptions must be made to obtain a solution. These assumptions are often imposed most conveniently by moving the problem into a filter space. We define filters $\{f_\gamma\}_{\gamma = 1}^L$ and set $y_\gamma = f_\gamma * y$ and $x^\tru_\gamma = f_\gamma * x^\tru$, so that
\beq
	y_\gamma = k^\tru * x_\gamma^\tru + n_\gamma
\eeq
for $\gamma \in [L] = \{1,\ldots,L\}$. Defining $\xf^\tru = \{x^\tru_\gamma\}_{\gamma=1}^L$, $\yf = \{y_\gamma\}_{\gamma=1}^L$, and $(k * \xf^\tru)_\gamma = k * x^\tru_\gamma$, we can write the filter space problem compactly as
\beq
	\yf = k^\tru * \xf^\tru + \nf.
	\label{model}
\eeq
The simplest nontrivial filter space is gradient space, where $L = 2$ and $f_1 = [1,-1], f_2 = [1,-1]^\Tr$, but there are many other possibilities. Determining $x$ from a filter space representation $\xf$ often does not work well, so typically one obtains an estimate $k$ of $k^\tru$ and deconvolves $y$ with $k$ to get $x$ \cite{levin2009understanding}.
%and solve for $k$ and $\xf = \{x_\gamma\}$ using the filtered data $\yf = \{y_\gamma\}$. After this, $x$ can be determined by non-blind deconvolution using $k$.

%Most methods for regularizing blind deconvolution fit into a Bayesian framework.most blind deconvolution methods do not fit perfectly, Bayesian inference is a convenient framework for articulating .
Bayesian inference is a convenient framework for imposing prior assumptions to regularize blind deconvolution \cite{campisi2007blind}. By assuming some distribution of $\nf$ we obtain a likelihood function $p(\yf \sepr k * \xf)$ which gives the probability that the data $\yf$ arose from a given pair $(k,\xf)$. We then choose priors $p(k)$ and $p(\xf)$ and compute the posterior distribution %of $(k,\xf)$ by Bayes' rule: 
\beq
	p(k, \xf \sepr \yf) \propto p(\yf \sepr k * \xf)p(\xf)p(k).
	\label{posterior}
\eeq
Estimates of $\xf$ and $k$ may be obtained by summary statistics on $p(k, \xf \sepr \yf)$. We call the mode of $p(k, \xf \sepr \yf)$ the joint maximum \emph{a posteriori} (MAP) estimator, while the mode of the marginal $p(k \sepr \yf) = \int p(k,\xf \sepr \yf) d\xf$ is the kernel MAP estimator. Most blind deconvolution methods are nominally MAP estimators but do not actually find a global minimizer, as this is typically intractable and may even be counterproductive. We refer to any method organized around optimizing a posterior as a MAP method, while methods that actually find a global minimum will be called \emph{ideal} MAP methods. 
%may be obtained by optimizing $- \log p(k \sepr \yf)$. This is usually computationally intractable because of the high-dimensional integral over $\xf$, but it can be
Joint MAP methods typically attempt to minimize the cost function $F(k,\xf) = - \log p(k, \xf \sepr \yf)$, which may be written (up to an irrelevant additive constant) as the sum of a data misfit and two regularization terms,
\begin{equation}
\begin{split}
	F(k,\xf) 	%&= - \log p(\yf \sepr k * \xf) - \log p(\xf) - \log p(k) \\
			&= L(k * \xf) + R_\xf(\xf) + R_k(k),
	\label{mapObj}
\end{split}
\end{equation}
where each of these functions may take the value $+\infty$ to represent a hard constraint. Kernel MAP estimation is more difficult as it involves a high-dimensional marginalization, and it is typically approximated by variational Bayes or MCMC sampling \cite{bishop2006pattern}.
Joint MAP estimation is the oldest, simplest, and most versatile approach to blind deconvolution \cite{ayers1988iterative, you1996regularization, chan1998total}, but initial joint MAP efforts on the camera shake problem met with failure \cite{fergus2006removing}, even when $\ell_p$ regularizers for $p < 1$ were used.  In \cite{levin2009understanding}, Levin \etal\, showed that the $\ell_p$ regularizer generally prefers blurry images to sharp ones: $\norm{\yf}_p^p < \norm{\xf^\tru}_p^p$, so that ideal joint MAP typically gives the trivial \emph{no-blur} solution $(k,x) = (\delta_0,y)$, where $\delta_0$ is the Kronecker delta kernel. The non-ideal joint MAP methods \cite{shan2008high, cho2009fast} somewhat compensate for the defects of ideal joint MAP by dynamic edge prediction and likelihood weighting, but benchmarking in \cite{levin2009understanding,levin2011marginal} showed that these heuristics sometimes fail.
%traced this failure to a lack of discriminatory power in the MAP cost function with $\ell_p$ regularizer

%a design flaw in the ideal joint MAP estimator. They reported that while hyper-Laplacian priors provide a good model of edges in natural images, using such priors in an ideal joint MAP estimator generally does not yield a sharp image. The problem is that the $\ell_p$ regularizer $R_\xf(\xf) = \norm{\xf}_p^p$ associated to a hyper-Laplacian prior generally prefers blurry images to sharp ones: $\norm{\yf}_p^p < \norm{\xf^\tru}_p^p$, so that ideal joint MAP typically gives the trivial \emph{no-blur} solution $(k,x) = (\delta_0,y)$, where $\delta_0$ is the Kronecker delta kernel. The joint MAP approaches of \cite{shan2008high, cho2009fast} somewhat compensate for this by dynamic edge prediction and likelihood weighting, but benchmarking in \cite{levin2009understanding,levin2011marginal} showed that these heuristics sometimes fail.
%the differences between joint and kernel MAP estimators were explained, and

In \cite{fergus2006removing} Fergus \etal\, developed a kernel MAP method with a sparse edge prior which was very effective for correcting camera shake. In  \cite{levin2009understanding} it was noted that marginalization over $\xf$ seems to immunize ideal kernel MAP against the blur-favoring prior problem. More refined kernel MAP methods were recently reported in \cite{levin2011marginal} and \cite{babacan2012bayesian}, and to our knowledge these two methods are the top performers on the benchmark 32 image test set from \cite{levin2009understanding}. While these efforts have made kernel MAP much more tractable, it remains harder to understand and generalize than joint MAP, so it would be useful to find a joint MAP method that is competitive with kernel MAP on the camera shake problem. % [cite: levin2011marginal, babacan].

In \cite{krishnan2011blind}, Krishnan \etal\, addressed the blur-favoring prior problem in joint MAP by changing the prior, proposing the scale-invariant \ratnorms ratio as a `normalized' sparse edge penalty. The $\ell_2$ normalization compensates for the way that blur reduces total $\ell_1$ edge mass, causing the \ratnorms penalty to prefer sharp images and  eliminating the need for additional heuristics. While their algorithm does not quite match the performance of \cite{fergus2006removing} on the benchmark test set from \cite{levin2009understanding}, it comes fairly close while being significantly simpler, faster, and in some cases more robust. Other promising joint MAP methods include \cite{cai2012framelet,wang2009multi,xu2010two}, but we are not aware of public code with full benchmark results for these methods. 

\subsection{Our approach}

\begin{figure*}%[h]
\centering
\includegraphics[scale=1.95]{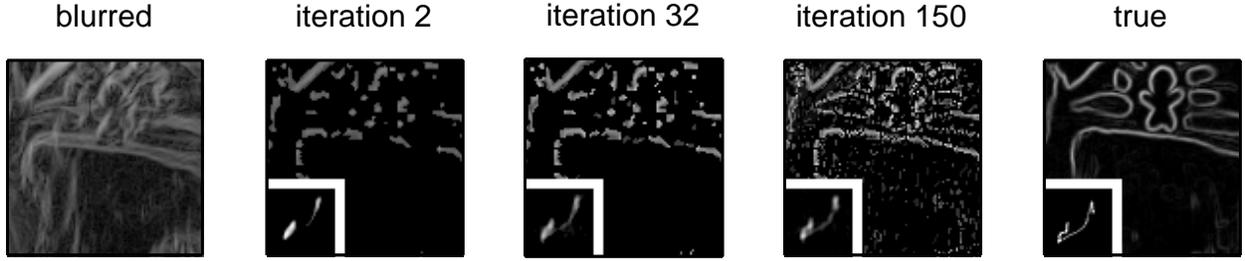}
\caption{Kernel estimation on an image from the test set of \cite{levin2009understanding}; a small patch has been selected and rescaled for clarity. \emph{Left:} Blurry edge map $|\yf|$. \emph{Center left to center right:} Evolution of the kernel $k$ (inset) and edge map magnitude $|\xf|$ in the final full-resolution stage. As $\tau$ increases in \eqref{map_sparse}, the edge map becomes less sparse and the kernel is refined. \emph{Right:}  $k^\tru$ and $|\xf^\tru|$. }
\label{alg_evol}
\end{figure*}

We propose a new approach to joint MAP blind deconvolution in which the kernel is estimated from a sparse approximation $\xf$ of the sharp gradient map $\xf^\tru$. Initially we constrain $\xf$ to be very sparse, so it contains only the few strongest edges in the image, and we determine $k$ such that $k * \xf \approx \yf$. Because $\xf$ is so much sparser than $\yf$, the solution $k = \delta_0$ is very unlikely. But generally this initial $k$ overestimates $k^\tru$, so we refine $k$ by letting weaker edges into $\xf$.
%obtained by this procedure will never be the incorrect delta kernel seen in ideal joint MAP

% can only be satisfied with a nontrivial $k$, so we avoid the incorrect delta kernel seen in ideal joint MAP. But our initial $k$ overestimates $k^\tru$, so we refine $k$ by iteratively letting weaker and weaker edges into $\xf$.

To present this approach formally, we set $f_1 = [1,-1], f_2 = [1,-1]^\Tr$, so that $\xf(p) = (x_1(p),x_2(p))$ is the discrete image gradient vector at each pixel $p$. We set $L(k,\xf) =  \tfrac{1}{2} \norm{k * \xf - \yf}^2_2$ and impose the usual positivity and unit sum constraints on $k$. We measure gradient sparsity using the $\ell_{2,0}$ norm:  $|\xf(p)|$ is the $\ell_2$ length of $\xf(p)$ and $\norm{\xf}_{2,0} = \norm{|\xf|}_0$ the number of nonzero gradient vectors. The joint MAP optimization problem is then
\begin{equation}
\begin{split}
	\minimize_{k, \xf} &\tfrac{1}{2} \norm{k * \xf - \yf}^2_2 \\
	\subto &k \geq 0, \,\, 1^\Tr k = 1, \,\, \norm{\xf}_{2,0} \leq \tau,
\label{map_sparse}
\end{split}
\end{equation}
where the expression $a^\Tr b$ denotes the dot product of the arrays $a$ and $b$ when considered as vectors, and the $1$ in $1^\Tr k$ is an all-ones array.

We solve this problem with an iterative optimizer described in \S \ref{algorithm}, and slowly increase $\tau$ as the iterations proceed. To initialize $\tau$ we use the \ratnorms ratio, a robust lower bound on a signal's sparsity \cite{lopes2012estimating}. The sharp $\xf$ should be significantly sparser than $\yf$, so initially we set $\tau = \beta_0 \tau_\yf$, where $\tau_\yf = \norm{|\yf|}_{1} / \norm{|\yf|}_{2}$ and $\beta_0 < 1$ is a small constant. After an initial burn-in period of $I_b$ iterations we multiply $\tau$ by a constant growth factor $\gamma > 1$, an action we repeat every $I_s$ iterations thereafter. 

We use a standard multiscale seeding technique to accelerate the kernel estimation step \cite{fergus2006removing,krishnan2011blind}. We begin by solving \eqref{map_sparse} with a heavily downsampled $\yf$, giving a cheap, low-resolution approximation to $k$ and $\xf$. We then upsample this approximation and use it as an initial guess to solve \eqref{map_sparse} with a higher resolution $\yf$, repeating the upsample-and-seed cycle until we reach the full resolution $\yf$. At each scale we use the same $\tau$ increase schedule. After kernel estimation we use non-blind deconvolution of $y$ with $k$ to get the sharp image $x$. 

The easiest way to understand how our kernel estimation works is to watch $k$ and $\xf$ evolve as the iterations progress. In Fig.~\ref{alg_evol}, the state of $k$ and $\xf$ is shown at iterations $2, 32,$ and $150$ of the final full-resolution scale, with $k^\tru$ and $\xf^\tru$ at far right. Initially $\xf$ is quite sparse, so $k$ cannot be a trivial kernel because the parts of $\yf$ not in $\xf$ must be attributed to blur. But this initial approximation is crude, so as $\tau$ increases with iteration, $\xf$ is allowed to have more and more edges so that $k$ can be refined. 

\subsection{Novelty and relations with existing methods}

%We do not know of any blind deconvolution methods that optimize with an $\ell_0$ constraint or penalty; the nearest approximations are the \ratnorms ratio used in \cite{krishnan2011blind} and the log prior used in \cite{babacan2012bayesian}.
Direct $\ell_0$ optimization is well-established in the compressed sensing community \cite{blumensath2010normalized,garg2009gradient} but we are not aware of any effective $\ell_0$ approaches to blind deconvolution. In \cite{krishnan2011blind} the \ratnorms ratio was deliberately chosen over $\ell_0$ because while both have the desired scale invariance, the graph of \ratnorms is smoother and looks more `optimizable' than $\ell_0$. We contend that $\ell_0$ may be difficult to use as a cost function, but very effective as a constraint. Gradient and kernel thresholding are commonly used \cite{shan2008high, cho2009fast} and these can be interpreted as $\ell_0$ projections, but they are typically used as auxiliary heuristics, not as the central modeling idea.  
Our technique of slowly increasing the sparsity constraint $\tau$ is reminiscent of matching pursuit algorithms for sparse approximation \cite{tropp2007signal,needell2009cosamp}. It is also related to the likelihood reweighting technique of \cite{shan2008high}, which may be seen by considering the Lagrangian of \eqref{map_sparse}. However, our initialization strategy requires that we use the constrained formulation rather than the Lagrangian. %We have found the constrained formulation to be more stable than the popular Lagrangian.% is that it allows us to set sparsity levels directly % rather than through a Lagrange multiplier. %find it easier and more intuitive to set sparsity levels directly rather than through a Lagrange multiplier.
%Recasting \eqref{map_sparse} in a Lagrangian form also reveals the likelihood reweighting technique of \cite{shan2008high}, but we find it more natural to adjust the sparsity directly instead of a Lagrange multiplier.
% through the
%, which is known to help prevent $\ell_p$ regularized joint MAP methods from falling into the blurry solution \cite{levin2009understanding}. 

%To see the relation we form the Lagrangian for \eqref{map_sparse}:
%\begin{equation}
%\begin{split}
%	\minimize_{k, \xf} &\tfrac{1}{2} \norm{k * \xf - \yf}^2_2 + \tfrac{1}{\lambda} \norm{\xf}_{2,0} \\ 
%	\subto &k \geq 0, \,\, 1^\Tr k = 1.
%\end{split}
%\label{map_sparse_lag}
%\end{equation}
%Lagrange multiplier theory says that given $\tau$, there is a $\lambda$ such that problems \eqref{map_sparse} and \eqref{map_sparse_lag} have the same solution. One can move the weight to the likelihood by multiplying the cost function by $\lambda$, so increasing the likelihood weight is equivalent to relaxing the sparse constraint. We use the $\tau$ formulation because it enables precise, scale-invariant sparsity control. % and the scale-invariant, robust \ratnorms initialization.  %working with the constraint gives us precise, scale-invariant control over $\norm{\xf}_{2,0}$.

\section{Alternating Projected Gradient Method} \label{algorithm}
%To accelerate convergence we use a multiscale pyramid strategy. Blind deconvolutions upsampled to obtain good initial guesses for more expensive fine-scale ones [cite: whoever]. Upsampling is performed by bilinear interpolation. At the coarsest scale the corresponding kernel is only $3 \times 3$, but we scale up to the full kernel after $S = 6$ stages.

%In each outer iteration we perform at most $I_\xf$ updates to $\xf$ and at most $I_k$ updates to $k$. 
%Each inner minimization subproblem is solved approximately by a few iterations of a projected gradient method, so each outer iteratio

To solve problem \eqref{map_sparse} at a given scale, we use a standard alternating descent strategy: starting from some initial $k$ and $\xf$, we reduce the cost function by updating $\xf$ with $k$ fixed, then $k$ with $\xf$ fixed, cycling until a stopping criterion is met. Each cycle, or outer iteration, consists of $I_\xf$ inner iterations updating $\xf$ and $I_k$ inner iterations updating $k$. All updates are computed with a projected gradient method; given a smooth function $h(u)$ and a constraint set $\USet$, projected gradient methods seek a solution of $\min_{u \in \USet} h(u)$ by updates of the form $u \str \Proj_\USet (u - \alpha_u g_u)$, where $g_u = \nabla h(u)$, $\alpha_u$ is a step size, and $\Proj_\USet(w)= \argmin_{u \in \SSet} \norm{u - w}_2^2$ is the Euclidean projection of $w$ onto $\USet$. Convergence of alternating descent and projected gradient methods to stationary points is proven in \cite{attouch2011convergence} under mild conditions. %While we do not verify the conditions here, the theory suggests that the empirical robustness of our algorithm should extend to general application. %helps to explain our algorithm's robustness.

We now describe how we compute the projected gradient iterations for the inner subproblems $\min_{k \in \KSet} L(k,\xf)$ and $\min_{\xf \in \XSet} L(k,\xf)$, where $L(k,\xf) = \tfrac{1}{2} \norm{k * \xf - \yf}^2_2$, $\KSet = \{ k \sepr k \geq 0, 1^\Tr k = 1\}$, and $\XSet = \{ \xf \sepr \norm{\xf}_{2,0} \leq \tau \}$. Letting $\rf = k * \xf - \yf$ denote the residual, we have % $\nabla_k L$ is given in terms of the residual $\rf = k * \xf - \yf$ as
$\nabla_k L  = \sum_\gamma \bar{\xf}_\gamma * \rf_\gamma$ and	 $\nabla_\xf L	= \bar{k} * \rf$,
%\begin{align}
%\nabla_k L  = \sum \bar{\xf}_\gamma * \rf_\gamma	 \quad \nabla_\xf L	= \bar{k} * \rf,
%\end{align}
where the bar denotes $180^\circ$ rotation about the origin. Assuming the nonzero elements of $|\xf|$ are distinct, the projection $\Proj_\XSet(\xf)$ is the top-$\tau$ vector thresholding
\beq
	P_\XSet(\xf)(i) = \xf(i) \cdot \mathbf{1}\left(|\xf(i)| \geq \theta(|\xf|,\tau)\right),
\eeq
where $\mathbf{1}(A)$ is the indicator function for condition $A$ and $\theta(|\xf|,\tau)$ is the $\tau^{th}$ biggest element of $|\xf|$. The set $\KSet$ is a canonical simplex with projection $\Proj_\KSet(k)$ given by
\beq
	P_\KSet(k)(i) = \max(0,k(i) - \sigma),
\eeq 
%$\sum_i P_\KSet(k)(i) = 1$.
where $\sigma$ is the unique solution of $1^\Tr P_\KSet(k) = 1$. Both $P_\XSet$ and $P_\KSet$ can be computed in linear time using selection algorithms \cite{cormen2009introduction, duchi2008efficient}.

 % is the point $u \in \SSet$ closest to $v$ in Euclidean distance:

%$\min_{u \in \USet} h(u)$, the projected gradient method
%Here $\Proj_\KSet, \Proj_\XSet$ are the Euclidean projections onto 
%The projected gradient method is essentially the steepest descent method augmented with Euclidean projections to enf
%This method

% to minimize the cost function by steepest descent step directions are used to minimize the cost function, while Euclidean projections enfroce

%using updates of the form
%Each cycle involves multiple projected gradient  %  $k$ and $\xf$ updates are of the form
%\beq
%	k \str \Proj_\KSet(k - \alpha_k \nabla_k L) \quad \xf \str \Proj_\XSet(\xf - \alpha_\xf \nabla_\xf L).
%\eeq
%and $\alpha_k, \alpha_\xf$ are step sizes. 

The step sizes $\alpha_\xf, \alpha_k$ are chosen by backtracking line search from an initial guess. In the $\xf$ subproblem our initial guess is 
%$\alpha_\xf = ()^\Tr (k * g_\xf)/ (g_\xf^\Tr g_\xf)$, where 
\beq
	\alpha_\xf = \frac{ (k * g_\xf)^\Tr \rf}{(k * g_\xf)^\Tr (k * g_\xf)},
\eeq   %For $\xf$ we compute the optimal step size
which is optimal in the sense that it solves the problem $\min_\alpha L(k, \xf - \alpha g_\xf)$. This aggressive step size was chosen over several alternatives, as it was the most effective for securing good edge support estimates. In the $k$ subproblem we use the spectral projected gradient (SPG) method \cite{birgin2000nonmonotone}; in the first iteration $\alpha_k = 1$, and in subsequent iterations we use the Barzilai-Borwein step size
\beq
	\alpha_k = \frac{(g_k - g_k^{old})^\Tr (g_k - g_k^{old})}{(g_k - g_k^{old})^\Tr (k - k^{old})}
\eeq
where $g_k^{old}$ and $k^{old}$ denote the values of $g_k$ and $k$ at the previous SPG iteration.%, and in the first iteration $\alpha_k  = 1$.
\begin{figure*}%[H]
\centering
%\begin{tabular}{c c}
%\includegraphics[scale=0.7]{err_cum_hist.eps}
\includegraphics[scale=0.85]{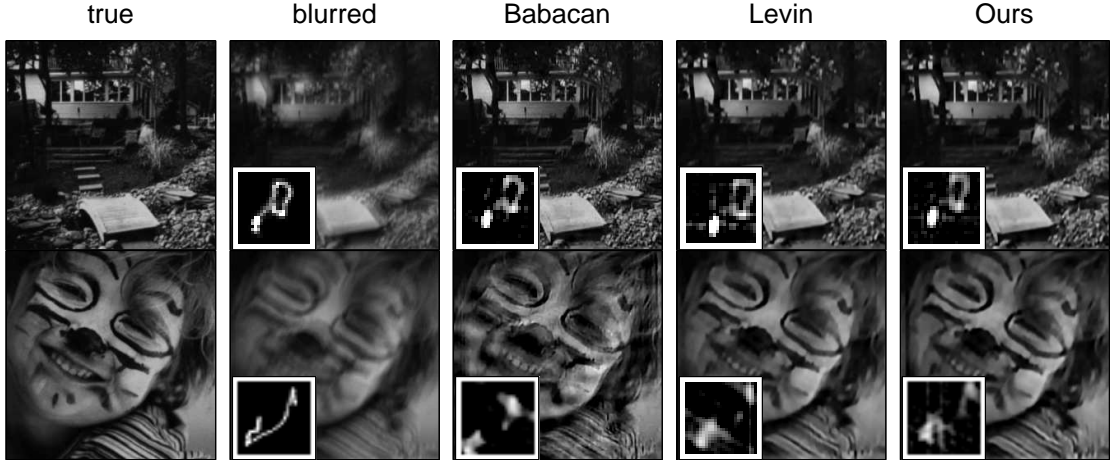} %& \includegraphics[scale=0.6]{err_cum_hist.eps}
%\end{tabular}
\caption{Sample results from our method, \cite{babacan2012bayesian,levin2011marginal} on the benchmark set of \cite{levin2009understanding}. True and recovered kernels inset.}
\label{sample_results}
\end{figure*}

%$S = 6$ we scale up to the full size kernel and image
% This is repeated until $k$ and $\xf$ reach their full size. [todo: how is the interpolation factor decided? maximum of 6 stages?]
%, and we solve problem \eqref{map_sparse} at this coarse scale. We then upsample $k$ and $\xf$ using bilinear interpolation [todo: by what factor?] and use them to initialize solution of \eqref{map_sparse} at the next scale.

\section{Implementation and Experiments} \label{experiments}

%Our algorithm's performance first experiment
%\begin{figure}
%\includegraphics[scale=0.25]{mukta.jpg}
%\end{figure}

We implemented our method in MATLAB by modifying the code of \cite{krishnan2011blind}, which uses a similar strategy of alternating minimization with multiscale seeding. The full-resolution kernel size was set to $35 \times 35$ for all experiments. The initial stage of the multiscale algorithm downsamples $\yf$ by a factor of $5/35$ in each direction, so that the kernel is of size $5 \times 5$, and each upsample cycle increases the size of $k, \xf$, and $\yf$ by a factor of roughly $\sqrt{2}$ until full resolution is reached. The parameters of the core single-scale algorithm from \S \ref{algorithm} were set to $\beta_0 = 0.15$, $\gamma = 1.10$, $I_b = 20$, $I_s = 10$, $I_\xf = 1$, $I_k = 6$. We do 30 iterations of the alternating projected gradient method for all scales except the final, full-resolution scale, which uses 180 iterations. Non-blind deconvolution with the estimated kernel was performed using the method of \cite{levin2007deconvolution}, using the parameter settings chosen in the code for \cite{levin2011marginal}. 

In \cite{levin2009understanding} a test set of 32 blurry images with known ground truth was created for benchmarking blind deconvolution methods. Each blurry image was formed by taking a picture of a sharp image with a camera that shook in-plane, and bright points outside the image were used to obtain ground truth blur kernels. A total of 32 blurry images were formed by blurring 4 sharp images on 8 different shake trajectories. This test set has become the \emph{de facto} standard for objectively comparing different methods.  

We ran our algorithm on this test set and compared its performance against the methods of \cite{levin2011marginal} and \cite{babacan2012bayesian}. We compare against these methods because they have published implementations which match or exceed the performance of the state-of-the-art methods in \cite{shan2008high, cho2009fast, krishnan2011blind,fergus2006removing}, and we know of no methods that outperform \cite{levin2011marginal} and \cite{babacan2012bayesian} on this test. We use the squared error metric $\text{SSE}(x) = \sum_i (x(i) - x^\tru(i) )^2$ to measure performance and note that results using the ratio metric of \cite{levin2009understanding} are similar. Results for \cite{levin2011marginal} were taken from files included with their published implementation, while results for \cite{babacan2012bayesian} were generated by running their online test script using the log prior, which was the best in their experiments. %o ensure fairness we restrict 

Our experiments were performed in MATLAB 2011b on an Intel Quad Core Xeon $2.2$ GHz Mac Pro. Our method's kernel estimation step took $45 - 60$ seconds per image, and deconvolution took 15 seconds. The other methods took $45 - 240$ seconds for kernel estimation, and their computation time depended strongly on kernel size. The difference is mostly due to our use of cheap SPG iteration rather than quadratic programming in the $k$ step, and also because $\Proj_\KSet$ and $\Proj_\XSet$ make $k$ and $\xf$ sparse, enabling $k * \xf$ to be computed faster.%iterates generated by our algorithm can be convolved together enables MATLAB to speed up convolutions, which are the leading order computational expense. %These are conservative settings , and better performance is sometimes obtained by

\begin{figure}
\centering
\includegraphics[scale=0.60]{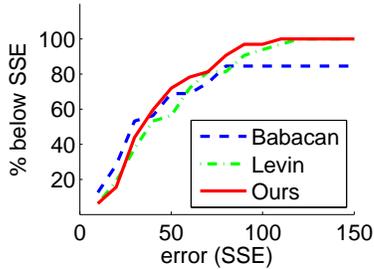}
\caption{Cumulative deblurring performance our method, \cite{levin2011marginal}, and \cite{babacan2012bayesian} on the 32 image test set of \cite{levin2009understanding}. The vertical axis is the percentage of the 32 runs having at most a given SSE.}
\label{sse_hist}
\end{figure}

%Cumulative SSE histograms for blind deconvolutions of image by our method, \cite{levin2007deconvolution}, and \cite{babacan2012bayesian} on the test set of . 
Sample results on the benchmark are shown in Fig.~\ref{sample_results}. Our method and \cite{levin2011marginal} perform very similarly on both images. On the house image \cite{babacan2012bayesian} is very sharp and by far the best, but it suffers from severe artifacts on the boy image.  Fig.~\ref{sse_hist} summarizes the full-benchmark performance of the three methods using cumulative error histograms. The curves for our method and \cite{levin2011marginal} are largely similar. The curve for \cite{babacan2012bayesian} is above ours and \cite{levin2011marginal} for about half of the images, but it flattens out below $85\%$ while the others plateau at $100\%$. This is because method \cite{babacan2012bayesian} struggled on the boy images. We note that the results reported in \cite{babacan2012bayesian} for this test set are better than those obtained in our run of their code, although we ran it without any modification. The authors of \cite{babacan2012bayesian} state that their code is a simplification of what was used to generate the reported results. While there may be a more sophisticated version of their code that outperforms ours, our method competes with the available state of the art.

%Our method is also significantly faster than \cite{levin2011marginal} and \cite{babacan2012bayesian}. While these typically take between $70 - 260$ seconds  to process an image, ours takes between $40 - 50$ seconds to estimate the kernel, and another $15$ seconds to deconvolve it. In many cases the kernel estimate is quite good after only $10 - 20$ seconds, as we saw in Fig.~\ref{example_bdc}. We ascribe the speed of our method to two factors. First, the iterations are cheaper than other methods; for example, the SPG iteration is cheaper than forming and solving a quadratic program in the $k$ step. Second, the hard thresholding and simplex projections on $k$ and $\xf$ keep them quite sparse, and MATLAB takes advantage of this to speed up convolutions. As convolutions are by far the most expensive operation in most deconvolution algorithms, one gets a significant speedup by carrying them out faster.

\section{Conclusion}

%In \cite{levin2009understanding}, Levin \etal\, showed that ideal joint MAP estimators do not generally fix camera shake in natural images when typical $\ell_p$ gradient priors are used. This left open the question of whether a different prior or a less strict joint MAP method could still do the job simply, quickly, and robustly. This question is important because the alternative kernel MAP methods proposed in \cite{fergus2006removing,babacan2012bayesian,levin2011marginal}, while very effective, are not as easy to generalize as joint MAP. %In \cite{krishnan2011blind} a promising joint MAP approach based on a \ratnorms edge penalty was proposed, but the optimization algorithm used was somewhat heuristic and over most of the test images from \cite{levin2009understanding} it did not perform as well as \cite{fergus2006removing}.% on the test set of \cite{levin2009understanding}. refinements in \cite{levin2011marginal} and \cite{babacan2012bayesian}.
%kernel MAP methods.
%a gradually relaxed $\ell_0$ penalty
%learning the kernel from the strongest edges in the image, then  incremental sparse approximation of  in gradient space.
% adding more and more edges using an incrementally relaxed $\ell_0$ constraint. 
We have proposed a blind deconvolution method in which the blur kernel is estimated by incremental sparse edge approximation. A rough global blur kernel is first estimated from only the strongest edges in the image, then it is refined as we allow the image edge map to gradually become less and less sparse.  Ours is the first simple, fast joint MAP method to match the state-of-the-art kernel MAP methods in \cite{levin2011marginal,babacan2012bayesian} on an objective benchmark. The success of the methods in \cite{krishnan2011blind} and this paper suggest that the downsides of ideal joint MAP described in \cite{levin2009understanding} can be robustly avoided without resort to a more complex kernel MAP estimation. %This is significant because joint MAP methods are generally simpler, faster, and easier to generalize than kernel MAP methods. %Along with \cite{krishnan2011blind}, our work 

There are many potential avenues for improving and extending our method. The edge sparsity relaxation schedule we use is slow and conservative, and a more adaptive schedule could make the method faster. Our initialization of the edge map sparsity does not take noise into account, and may need to be modified for very noisy images. Extension to nonuniform blur models, nonquadratic likelihoods, and fast parallel or GPU implementations are possible. The speed of our kernel estimation may make it useful as an input to high-quality non-blind methods such as \cite{babacan2012bayesian}.

%One interesting feature of our algorithm is it can find rough kernel approximations quite quickly. For example the kernel in the center left panel of Fig.~\ref{alg_evol} is found after only about 10 seconds. 
%A GPU implementation for fast operation on large images would also be of interest. % set the incremental sparsity schedule, in particular the factors $\beta$ and $\gamma$

% convolution, projection, and upsampling operations,    %that follows the joint MAP principle less strictly.  %quasi-MAP method.  %join with \cite{krishnan2011blind} in arguing that

%, it differs from the implementation used to generate their published results. % the available code and paper lacks sufficient detail to reproduce them. In any case 

%results on this test set reported in \cite{babacan2012bayesian} are better than these, but we cannot account for the discrepancy because the published code differs in unspecified ways from that used to generate their results.

%Better results than these are claimed in \cite{babacan2012bayesian} for their method. .   %the horizontal coordinate is SSE value, and the vertical coordinate is the the percentage of blind deconvolutions whose SSE is below the specified threshold. curve height is the percentage of 

\newpage

\bibliographystyle{IEEEbib}
\small{\bibliography{/Users/shearerp/Desktop/master_bib/master_bib_v2}}

\end{document}